\documentclass[letterpaper]{article} 
\usepackage{aaai2026}  
\usepackage{times}  
\usepackage{helvet}  
\usepackage{courier}  
\usepackage[hyphens]{url}  
\usepackage{graphicx} 
\usepackage{natbib}  
\usepackage{caption} 
\usepackage{algorithm}
\usepackage{algorithmic}
\usepackage{amsmath}
\usepackage{amssymb}
\usepackage{appendix}
\usepackage{multirow}
\usepackage{subcaption}
\usepackage{makecell}
\usepackage{booktabs}
\usepackage{xcolor}

\usepackage{newfloat}
\usepackage{listings}
\DeclareCaptionStyle{ruled}{labelfont=normalfont,labelsep=colon,strut=off} 
\floatstyle{ruled}
\newfloat{listing}{tb}{lst}{}
\floatname{listing}{Listing}
\newcounter{checksubsection}
\newcounter{checkitem}[checksubsection]

\title{OFL-SAM2: Prompt SAM2 with Online Few-shot Learner \\ for Efficient Medical Image Segmentation}

\author {
    Meng Lan\textsuperscript{\rm 1},
    Lefei Zhang\textsuperscript{\rm 2},
    Xiaomeng Li\textsuperscript{\rm 1}\thanks{Corresponding author.}
}
\affiliations {
    \textsuperscript{\rm 1} Department of Electronic and Computer Engineering, The Hong Kong University of Science and Technology\\
    \textsuperscript{\rm 2}National Engineering Research Center for Multimedia Software, School of Computer Science, Wuhan University \\
     \{eemenglan, eexmli\}@ust.hk, zhanglefei@whu.edu.cn
}

\usepackage{bibentry}

\begin{document}

\maketitle

\begin{abstract}
The Segment Anything Model 2 (SAM2) has demonstrated remarkable promptable visual segmentation capabilities in video data, showing potential for extension to medical image segmentation (MIS) tasks involving 3D volumes and temporally correlated 2D image sequences. However, adapting SAM2 to MIS presents several challenges, including the need for extensive annotated medical data for fine-tuning and high-quality manual prompts, which are both labor-intensive and require intervention from medical experts. To address these challenges, we introduce OFL-SAM2, a prompt-free SAM2 framework for label-efficient MIS. Our core idea is to leverage limited annotated samples to train a lightweight mapping network that captures medical knowledge and transforms generic image features into target features, thereby providing additional discriminative target representations for each frame and eliminating the need for manual prompts. Crucially, the mapping network supports online parameter update during inference, enhancing the model’s generalization across test sequences. Technically, we introduce two key components: (1) an online few-shot learner that trains the mapping network to generate target features using limited data, and (2) an adaptive fusion module that dynamically integrates the target features with the memory-attention features generated by frozen SAM2, leading to accurate and robust target representation. Extensive experiments on three diverse MIS datasets demonstrate that OFL-SAM2 achieves state-of-the-art performance with limited training data. Code will be released at https://github.com/xmed-lab/OFL-SAM2.
\end{abstract}

\section{Introduction}
Medical Image Segmentation (MIS) is an important step in medical image analysis, as it can assist in downstream applications such as disease diagnosis and monitoring of disease progression \cite{li2018h,li2018semi,li2020transformation}. Recently, the Segment Anything Model (SAM) \cite{huai2025tmi,zhang2024glandsam} has gained significant attention for its powerful segmentation abilities and prompt-based interactions. Nonetheless, SAM's zero-shot performance in MIS is suboptimal due to the domain gap between natural images and medical images.

\begin{figure}[t]
  \centering
  \includegraphics[width=1.0\linewidth]{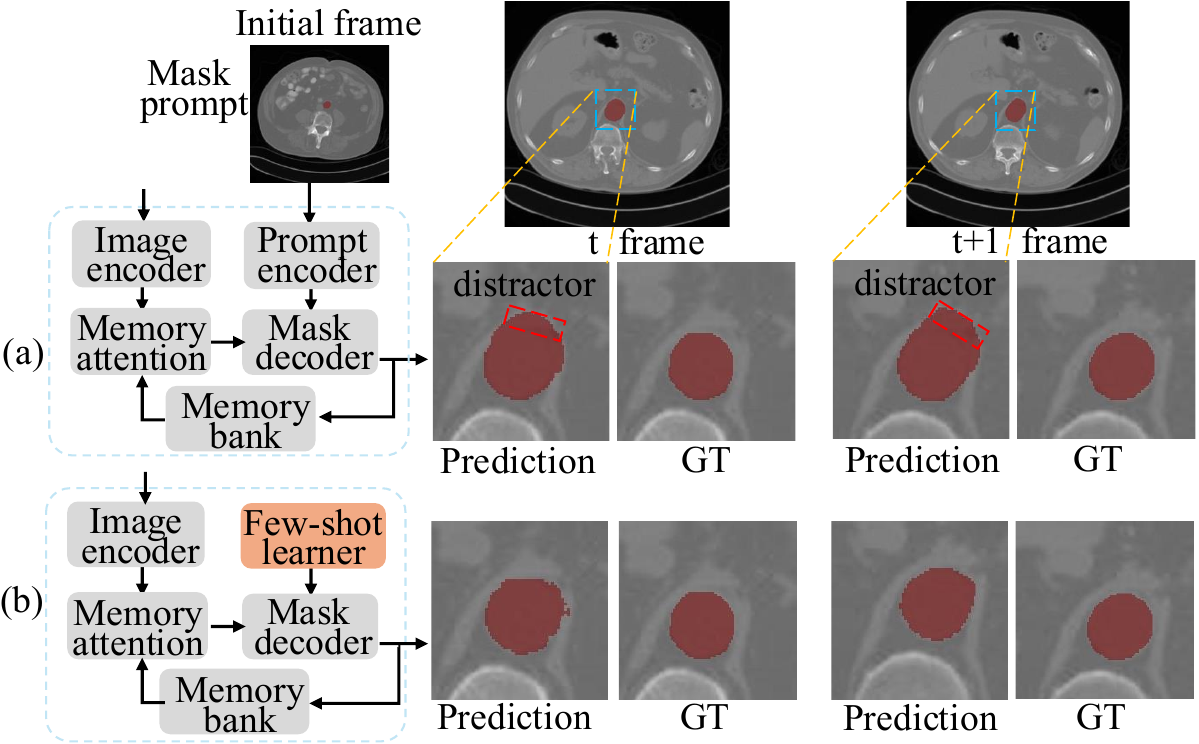}
  \caption{Comparisons of (a) SAM2 model and (b) our OFL-SAM2 model. SAM2 is susceptible to adjacent distractors.}
  \label{fig1}
\end{figure}

To bridge this gap, previous works primarily focus on full fine-tuning \cite{ma2024segment, zhu2024medical,wang2023sam} or parameter-efficient fine-tuning (PEFT) \cite{xiao2024cat,wang2024tri} using large annotated medical datasets. However, the fine-tuned medical SAM variants still require manual prompts for every frame, which requires sustained medical expert intervention, especially when processing 3D volumes as anatomically continuous 2D image sequences. SAM2 \cite{sam2} mitigates this issue through its streaming memory mechanism, enabling whole-sequence target segmentation with just a single-frame prompt, which has spurred interest in SAM2-based variants. Yet, these still rely on user-provided prompts, which hinders the automated processing of extensive images and is unfriendly to those without expertise. These limitations have driven the emergence of prompt-free SAM methods.

Current prompt-free medical SAM models typically replace manual prompts through three main strategies: (1) fine-tuning SAM models using PEFT strategies for direct semantic segmentation \cite{samed,hsam}, (2) employing prototype learning to acquire class-specific representative knowledge for self-prompt generation \cite{surgicalsam,yan2025pgp}, (3) generating coarse masks via registration or regression methods from training sets, which are then converted into sparse/dense prompts \cite{xu2024sam}. These prompt-free models process 3D volumes and temporally correlated 2D image sequences, e.g., surgical videos, in a frame-by-frame manner, providing a self-generated prompt for each frame, but they fail to leverage the spatiotemporal contextual information. In contrast, the SAM2 framework that effectively utilizes spatiotemporal features in sequential data shows promising research potential. One potential solution is to adapt existing prompt-free SAM methods to the SAM2 architecture. However, some of them, such as the leading H-SAM\cite{hsam}, are incompatible with SAM2, as the memory encoder designed for binary mask prediction cannot encode their multi-class semantic predictions into the memory bank to guide the segmentation, whereas other methods are either limited by performance or require extensive fine-tuning data. Researchers have also explored the direct adaptation of SAM2 to 3D MIS \cite{revsam2}. For instance, the recently proposed FATE-SAM2 \cite{he2025few} retrieves images similar to the query image from the training set and combines them with past inference frames to construct memory embeddings. However, the performance of these prompt-free SAM2 variants remains unsatisfactory and is even inferior to leading prompt-free SAM-based methods. Upon careful investigation, we found that the core issue lies in the SAM2 framework itself, i.e., relying solely on the pixel-level feature matching based memory attention in SAM2 can make the model struggle to discriminate adjacent distractors, a critical challenge in medical images, where ambiguous boundaries frequently create distractors around targets, as shown in Fig. \ref{fig1}. This may be attributable to the fact that SAM2 is unable to provide discriminative target representations generated by a prompt for each frame as the SAM does.

In this study, we propose OFL-SAM2, an online prompt-free SAM2 framework for label-efficient MIS, aimed at providing discriminative target features for each frame like the prompt features of the SAM model by training a mapping network that can transform generic image features into target features using limited annotated samples, while preserving the temporal contextual information of SAM2. Based on SAM2, we introduces two lightweight yet effective components: 1) an online few-shot learner that trains the mapping network using limited data and updates the network parameters online during inference, and (2) an Adaptive Fusion Module (AFM) that dynamically integrates the target features with the memory-attention features to adapt the frozen decoder of SAM2 and suppress the potential distractor representations. Given the limited frame-mask pairs in the training set, OFL-SAM2 utilizes the image and memory encoders of SAM2 to extract the generic image features and the target features, which serve as the input and supervision for the mapping network, respectively. The mapping network is then optimized by the few-shot learner. During inference, OFL-SAM2 processes the query image by simultaneously generating target features through the mapping network and memory-attention features via SAM2, with AFM dynamically fusing these features for the decoder while employing a quality-aware selection mechanism to update both the mapping network and memory bank. Notably, the entire adaptation requires modest computational overhead as the mapping network and AFM each consist of just a single convolutional layer while keeping the core SAM2 components completely frozen, enabling rapid training convergence with limited annotations. Comprehensive evaluations across three diverse MIS datasets confirm that OFL-SAM2 achieves state-of-the-art performance for label-efficient MIS.

The primary contributions of this study are as follows:
\begin{itemize}

\item We propose a novel prompt-free SAM2 framework for label-efficient MIS, where we design an online few-shot learner that effectively utilizes both limited training samples and incoming test samples to continuously train a lightweight mapping network, enabling it to capture domain-specific medical knowledge and generate discriminative target representations for each frame in the sequence without requiring manual prompts.

\item We develop an adaptive fusion module to dynamically integrate the learned target features with the memory-attention features to adapt the frozen SAM2 decoder and suppress the potential distractor representations.

\item We evaluate OFL-SAM2 on three MIS datasets with multiple modalities, including CT, MRI, and surgical videos. The results demonstrate that OFL-SAM2 achieves state-of-the-art performance with limited training data.

\end{itemize}

\section{Related Work}
\subsection{SAM-based Medical Image Segmentation}
The SAM \cite{sam2023}, trained on over a billion natural images, has demonstrated remarkable zero-shot segmentation capabilities when provided with visual prompts (e.g., a point or a bounding box). However, while SAM excels with natural images, its performance significantly degrades on medical images as evidenced by multiple studies \cite{deng2023sam, cheng2023sam, roy2023sam}. This performance gap has spurred considerable research into adapting SAM for medical images, primarily through various fine-tuning approaches  \cite{wu2025medical,lin2024beyond}. For instance, MedSAM \cite{ma2024segment} created a large-scale medical image dataset to retrain SAM with bounding box prompts; SAMed \cite{samed} incorporated LoRA \cite{hu2022lora} layers into the image encoder while retaining the original mask decoder; H-SAM \cite{hsam} enhanced medical feature extraction through LoRA-modified encoders and introduced a hierarchical prompt-free decoder. MSA \cite{wu2025medical} developed specialized spatial and depth adapters for 3D medical images.

Building upon these SAM adaptations, researchers have similarly sought to optimize SAM2 for MIS. For example, MedicalSAM2 \cite{zhu2024medical} and MedSAM2 \cite{ma2025medsam2} fine-tuned the image encoder and mask decoder of SAM2 using extensive medical data. SAM2-Adapter \cite{chen2024sam2} introduced lightweight adapters into the image encoder, which are fine-tuned together with the mask decoder. Some studies also explored the prompt-free SAM2 for MIS \cite{bai2024fs}. E.g., FATE-SAM2 \cite{he2025few} encoded some support image-mask pairs into the memory bank as memory features to launch the SAM2 inference directly. RevSAM2 \cite{revsam2} proposed the reverse propagation strategy to select high-quality query information for the memory bank. In this work, we advance the prompt-free adaptation of SAM2 for efficient MIS by introducing an online few-shot learner that generates discriminative target representations for each frame like a prompt while maintaining strong generalization across test sequences.

\begin{figure*}[t]
  \centering
  \includegraphics[width=0.86\linewidth]{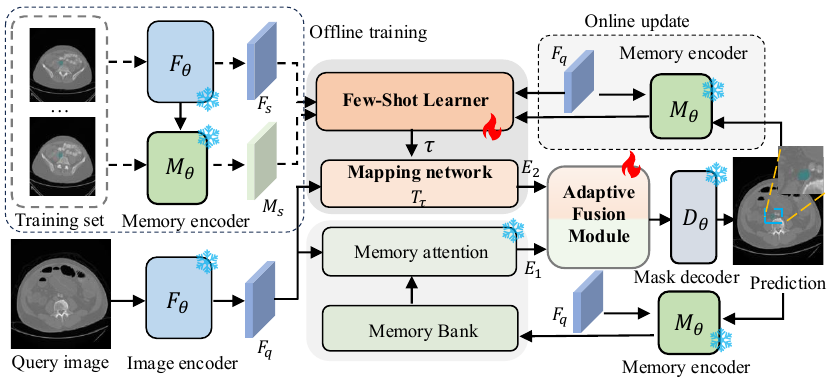}
  \caption{An overview of our OFL-SAM2 framework.}
  \label{framework}
\end{figure*}

\subsection{Online Discriminative Learning}
The concept of online discriminative learning initially gained prominence in visual object tracking \cite{wang2021transformer,danelljan2020}, where its combination of strong performance and computational efficiency made it particularly effective, with these early approaches employing optimized convolutional filters trained online to perform robust foreground-background classification across video frames. This methodology was subsequently extended to video object segmentation, as demonstrated by FRTM \cite{robinson2020} which incorporated Conjugate Gradient and Gauss-Newton optimization to enable its few-shot learner to construct target-specific models from minimal template data during inference. The framework was further refined by LWL \cite{lwl2020} through the introduction of a label encoder to generate information-rich few-shot labels, while JOINT \cite{mao2021joint} achieved complementary feature representation by combining transductive and online inductive features. More recently, this online learning paradigm has been successfully adapted to vision-language models, exemplified by Meta-Adapter \cite{song2023meta} which developed a lightweight residual-style adapter to enhance CLIP features under the guidance of a few-shot learner, showcasing the continued evolution and expanding applications of this approach across various domains.

\section{Method}

\subsection{Overall Pipeline}
Mathematically, we first divide a 3D volume into an anatomically continuous image sequence denoted as X, and its $i$th image is denoted as $x_i \in \mathbb{R}^{H \times W \times 3}$. Given the sequence of query images as $Q \in \mathbb{R}^{N \times H \times W \times 3}$, the training set $S =\{X_s, Y_s\}$, where $X_s = \{x_{1}, x_{2},...x_{n}\}$ represents the training images and $Y_s = \{y_{1}, y_{2},...y_{n}\}$ represents their corresponding labels, our goal is to predict the segmentation masks $P \in \mathbb{R}^{N \times H \times W}$ of $Q$ using the limited training image-label pairs $S$. Notably, the training set may consist of multiple different medical image sequences.

As illustrated in Fig. \ref{framework}, our OFL-SAM2 builds on SAM2 by removing the prompt encoder and including the designed few-shot learner and adaptive fusion module. OFL-SAM2 could be mainly divided into two branches: the online branch (the few-shot learner) and the offline branch (the memory attention module). In the training process, the training set is used to rapidly train the few-shot learner and the adaptive fusion module with the original modules of SAM2 frozen. During the inference, OFL-SAM2 sequentially processes the query images from sequence $Q$. The image encoder $F_{\theta}$ first extracts the generic features of the query image $Q_i$, which are sent to the two branches. The memory attention module produces the memory-conditioned feature $\mathbf{E}_{1}$, and the mapping network transforms the generic features into the target features $\mathbf{E}_{2}$. $\mathbf{E}_{1}$ and $\mathbf{E}_{2}$ are then intelligently fused through our adaptive fusion module to generate robust target features for the frozen decoder $D_{\theta}$, yielding accurate final predictions. A quality assessment mechanism evaluates each output mask prediction to determine its suitability for updating both the memory bank and the mapping network parameters.

\subsection{Memory Attention module}
This offline branch directly utilizes the off-the-shelf memory attention module of SAM2, which performs the core cross-attention operation between the query image features and memory features in the memory bank to achieve the target information propagation. Since there is no user-provided prompt, we need to store some reference features in the memory bank to launch the memory attention module. To this end, we first select two image-mask pairs from the training set $S$ that are most similar to the input query image $Q_i$. Given the image features $F_{Q_i} = F_{\theta}(Q_i)$ of $Q_i$ and the image features $F_{S_j} = F_{\theta}(S_j)$ of the image $x_{j}$ in training set $S$, we calculate the cosine similarity as follow:

\begin{equation}
\operatorname{Sim}\left(F_{Q_i}, F_{S_j}\right)=\frac{F_{Q_i} \cdot F_{S_j}}{\left\|F_{Q_i}\right\| \cdot\left\|F_{S_j}\right\|}
\end{equation}

Based on the similarity scores, a ranked list of training image features is generated. Then the top K=2 most similar training image features and their corresponding masks are sent into the memory encoder to produce the memory features, which are then stored in the memory bank. These selected training examples may come from different training sequences, providing diverse reference anatomical information for the segmentation process. 

The memory attention module contains four sequential memory attention layers, each layer comprises a self-attention module, a cross-attention module and a feed-forward network. In each attention layer, the query image features $F_{Q_i}$ first go through the self-attention layer and then are fed into the cross-attention layer along with the memory features to retrieve and get the target representation from the memory features. Finally, the output features are obtained through the residual connection and feed-forward network. After four iterations of the memory attention layers, the memory-conditioned target features $\mathbf{E}_{1}$ are obtained. After the adaptive fusion module and the mask decoder, OFL-SAM2 generates the mask prediction of the query image.

\subsection{Few-shot Learner}
In this module, an online few-shot learner $A_{\theta}$ is designed to predict the parameters $\tau$ of the mapping network $T_{\tau}$ by minimizing squared error on the training samples, which can be formulated as follows:
\begin{equation}
\begin{aligned}
	L(\tau)=\frac{1}{2} \sum\left\|T_{\tau}\left(F_{S_i}\right)- M_{S_i}\right\|^{2} +\frac{\lambda}{2}\|\tau\|^{2}.
\label{eq_3}
\end{aligned}
\end{equation}

Here, $F_{S_i}$ is the generic features of the image $x_{i}$ in the training set and $M_{S_i}$ is the corresponding memory features. The mapping network $T_{\tau}$, as a differentiable linear layer, where $\tau \in \mathbb{R}^{ K \times K \times C \times D}$ is the weight of a convolutional layer with kernel size K, maps the $C$-dimension generic image feature to the $D$-dimension target-aware representation with the spatial size unchanged. $\frac{\lambda}{2}\|\tau\|^{2}$ is the regularization item, and $\lambda$ is a learnable weight. The steepest descent method \cite{bhat2020learning} is applied to iteratively minimize the squared error and optimize the weight $\tau$. After the training of the few-shot learner, the trained mapping network is applied to the query image feature $F_{Q_i}$ to generate the target representation $\mathbf{E}_{2}$.
Before feeding into the subsequent adaptive fusion module, the channel dimension of $\mathbf{E}_{2}$ is converted from $D$ to $C$ by a convolutional layer to facilitate subsequent feature fusion. During inference, the high-quality mask prediction and its corresponding image features will be encoded as the training samples to update the mapping network parameters online through the few-shot learner.

\subsection{Adaptive Fusion Module}
Compared to the original SAM2, OFL-SAM2 introduces an additional feature flow generated by the mapping network. However, given that only a limited number of training samples are available, which are insufficient for effectively fine-tuning the decoder, we keep the decoder parameters frozen. In this case, how to effectively fuse the features of the two branches such that the data distribution of the fused features closely resembles that of the features input to the decoder in the original SAM2 is an urgent problem to be solved. At the same time, how to suppress the potential distractor representations in the features of the two branches, so that the fused features have a more robust and discriminative target representation is also a challenge to be considered. To address these issues, we propose the adaptive fusion module. Our AFM employs a weight network to learn an element-wise weight map $W$ for the features of the offline branch, while simultaneously applying a complementary weight map $(1-W)$ to the online branch features. This sophisticated weighting mechanism performs pixel-level feature recalibration, intelligently balancing the contributions from both branches and suppressing the irrelevant features to produce robust target features $E_{tar}$ that maintain the original SAM2's expected data distribution. The AFM could be mathematically described as follows:

\begin{equation}\label{AFM}
\begin{aligned}
	&W = G_\theta([E_1, E_2]),\\
	& E_{tar} = W \odot E_1  + (1-W) \odot E_2,
\end{aligned}
\end{equation}
where $G_\theta$ is the learnable weight network, $[ \cdot ]$ indicates the concatenation, and $\odot$ indicates element-wise multiplication. The weight network is implemented with a 3$\times$3 convolution layer followed by a sigmoid function.

\subsection{Update Strategy}

Different from the strategy of SAM2 that indiscriminately stores all the predicted mask and the corresponding image features into the memory bank, we select high-quality predictions to update the memory bank and the mapping network parameters to prevent incorrect predictions from accumulating in the memory bank and misleading the mapping network. We adopt the confidence score to assess the quality of the mask prediction. Let $P_{(i,j)}$ denotes the probability of the mask prediction at location $(i, j)$, and $\hat{G}$ represents the predicted binary mask, where $\hat{G}_{(i,j)}$ equals 1 when the probability of pixel at $(i, j)$ is higher than a threshold of 0.5 and the pixel is predicted as the target object, otherwise it equals 0. we filter out the background and calculate the confidence score of the mask prediction as follows:
\begin{equation}
\mathcal{S}_{cf}=\frac{\sum_{i, j} P_{i,j} \cdot \hat{G}_{i, j}}{\sum_{i, j} \hat{G}_{i, j}}
\end{equation}
If $\mathcal{S}_{cf}$ is higher than a threshold $\gamma$ = 0.8, the prediction $P$ and the corresponding query image features $F_{Q_i}$ are fed into the memory encoder to produce the memory features $M_{Q_i}$. Then, $F_{Q_i}$ and $M_{Q_i}$ are sent into the few-shot learner as the training sample to update the mapping network parameters online, and $M_{Q_i}$ is stored in the memory bank.

\begin{table*}[t]
  \centering
  \setlength{\tabcolsep}{1.3mm}
  \begin{tabular}{c|cccccccc|cc}
    \toprule
    \makecell{Method}&{Spleen}&\makecell{Right\\Kidney}&\makecell{Left\\Kidney}& \makecell{Gallb-\\ladder}&Liver&Stomach&Aorta&Pancreas& \makecell{Mean \\ Dice$\uparrow$(\%)}  &HD $\downarrow$\\
    \midrule
    SAM2 \cite{sam2} & 78.24 & 82.67 & 77.94 &66.42 & 75.59 & 72.35 & 68.76 & 48.31 & 71.29 & 24.08 \\ 
    \midrule
    \multicolumn{11}{c}{\makecell{Training set: 2 volumes}} \\
    \midrule
    SAMed \cite{samed} &84.62 & 80.70 &81.57 & 62.60 & 90.44 & 66.33 & 77.50 & 51.46 & 74.41 & 21.49 \\
    SurgicalSAM \cite{surgicalsam} & 87.03 & 85.31 & 86.72 & 70.44 & 85.18 & 64.77 & 85.46 & 47.98 & 76.61 &17.42 \\
    H-SAM \cite{hsam} &88.28 & 82.79 & 84.00 & 69.95 & \textbf{91.99} & \textbf{75.85} & 83.68 & 55.71 &79.03 &15.32 \\
    \midrule
    FS-MedSAM2 \cite{bai2024fs}& 85.37 & 89.14 & 88.30 & 68.67 & 80.88 & 57.91 & 85.78 & 45.27 & 75.16 & 20.28 \\
    FATE-SAM2 \cite{he2025few}& 85.49 & 89.59 & 88.61 & 68.59 & 81.60 & 58.55 & 86.01 & 45.28 & 75.47 & 19.10 \\
    \textbf{OFL-SAM2} & \textbf{90.11} & \textbf{90.43} & \textbf{89.62} & \textbf{72.64} & 90.52 & 71.59 & \textbf{88.45} & \textbf{61.17} &\textbf{81.81}&\textbf{9.10}\\
    \midrule
    \multicolumn{11}{c}{\makecell{Training set: 3 volumes}} \\
    \midrule
    SAMed \cite{samed} &86.07 & 82.40 &82.92 & 63.75 & 92.42 & 67.65 & 79.07 & 52.54 & 75.84 & 19.15 \\
    SurgicalSAM \cite{surgicalsam} & 87.86 & 86.19 & 87.38 & 71.39 & 85.95 & 65.50 & 86.29 & 48.86 & 77.43 &16.21 \\
    H-SAM \cite{hsam} &89.53 & 83.74 &85.10 & 71.40 & \textbf{93.59} & \textbf{76.65} & 84.99 & 56.86 
    &80.12 &13.77 \\
    \midrule
    FS-MedSAM2 \cite{bai2024fs} & 86.11  & 89.91 & 89.12 & 69.80 & 81.72 & 58.53 & 86.63 & 45.88 & 75.96 &18.90 \\
    FATE-SAM2 \cite{he2025few} & 86.34  & 90.64 & 89.56 & 69.69 & 82.35 & 59.25 & 87.16 & 46.08 & 76.38 &18.33 \\
    \textbf{OFL-SAM2} &\textbf{ 90.82} & \textbf{91.28} & \textbf{90.24} & \textbf{73.54} & 91.27 & 72.19 & \textbf{89.10} & \textbf{61.75} &\textbf{82.52}&\textbf{8.22}\\
    \hline
  \end{tabular}
  \caption{Comparison on Synapse-CT dataset. The Dice scores of the eight organs are reported. SAM2 adopts a mask prompt.}
  \label{tab:synapse}
\end{table*}

\section{Experiments}

\subsection{Dataset and Evaluation}
We conduct experiments on three medical datasets with different modalities, including Synapse-CT \cite{landman2015miccai}, PROMISE12~\cite{litjens2014evaluation} and Autolaparo \cite{autolaparo}. We utilize the Dice score (DSC) and the average Hausdorff distance (HD) as evaluation metrics.

The \textbf{Synapse-CT} multi-organ segmentation dataset has 30 cases in total and each CT volume contains 85 to 198 slices with a resolution of 512$\times$512. Following SAMed \cite{samed}, we evaluate eight abdominal organs. \textbf{PROMISE2012} dataset contains 50 3D transversal T2-weighted MR images of the prostate with manual binary prostate gland segmentation and is obtained from multiple centers with different acquisition protocols. The resolution of the PROMISE12 dataset is 512$\times$512. \textbf{Autolaparo} is a surgical instrument segmentation dataset, which contains 300 laparoscopic video sequences and each sequence has 6 consecutive frames. Autolaparo focuses on 4 types of instruments and provides 8 types of segmentation annotations, where the shaft and manipulator of each instrument are annotated separately.

During the training process, for the Synapse-CT and PROMISE12 datasets, we construct two types of training sets by randomly selecting 3 and 2 volumes of data, and treat each 3D volume as an anatomically continuous 2D image sequence. For the Autolaparo dataset, we construct two types of training sets by randomly selecting 10\% and 5\% of the whole surgical video sequences and treating them as temporally correlated 2D image sequences. The remaining data of each dataset is used for evaluation.

\subsection{Implementation details}
We train our model on one NVIDIA RTX 3090 GPU using the SAM2\_base model as our adapted SAM2 model. Considering the powerful segmentation capability of SAM2 and the limited labelled data, we chose to freeze the parameters of all SAM2 modules during the training process. The training loss is a combination of Cross-Entropy loss and Dice loss. The maximal training epoch is set to 40. The AdamW optimizer \cite{adamW} is employed for model optimization, with an initial learning rate of 1e-3. The learning rate decays by a factor of 0.1 at the 10th and 30th epochs. We conduct individual segmentation for each class in the datasets and report the average performance of three groups of random training data. In the few-show learner, $C$=256, $D$=64. Besides, during training, for each training sample, our few-shot learner employs 10 iterations of optimization to train the mapping network parameters $T_{\tau}$ with the default parameter initialization, while in the inference process, we reduce the number of iterations to 5.


\subsection{Comparison with State-of-the-art Methods}
To evaluate the performance of our proposed OFL-SAM2, we compared it with state-of-the-art methods on the three datasets. The comparison methods include the SAM2 model with mask prompt \cite{sam2}, the prompt-free SAM variants: SAMed \cite{samed}, SurgicalSAM \cite{surgicalsam}, and H-SAM \cite{hsam}, and the prompt-free SAM2 variants: FS-MedSAM2 \cite{bai2024fs} and FATE-SAM2 \cite{he2025few}.

\begin{figure*}
    \centering
    \includegraphics[width=0.92\linewidth]{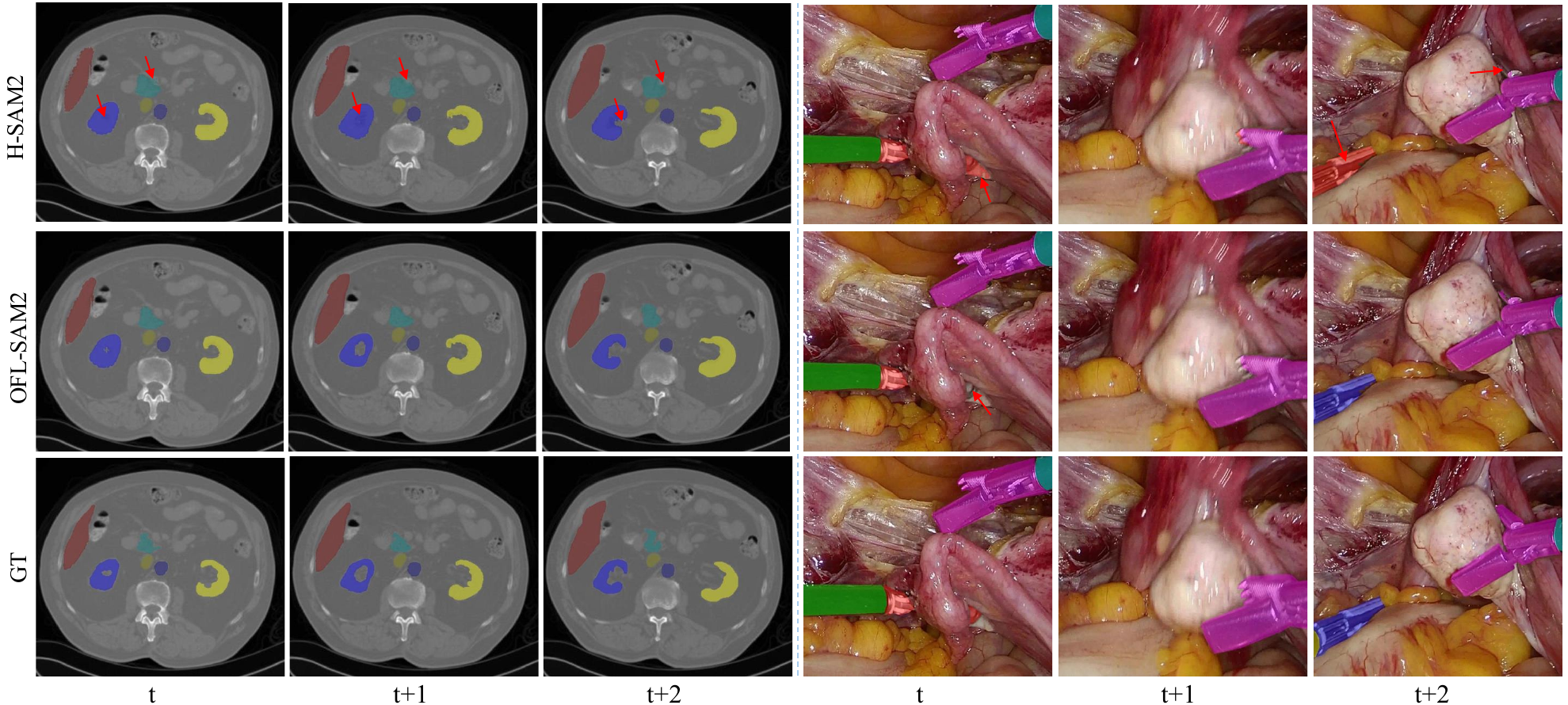}
    \caption{Visualization comparison between H-SAM and OFL-SAM2 on Synapse-CT (left) and Autolaparo (right) datasets.}
    \label{vis_res}
\end{figure*}

\noindent\textbf{Synapse-CT.} As shown in Table~\ref{tab:synapse}, OFL-SAM2 exhibits outstanding performance on the Synapse-CT dataset under different training conditions. When using 3 training volumes, our OFL-SAM2 achieves 82.52\% mean DSC and 8.22 HD, surpassing all competing methods by a significant margin, including a 6.14\% DSC improvement over the SAM2 variant FATE-SAM2. When reducing the training set to merely 2 volumes, OFL-SAM2 maintains its performance advantage while exhibiting remarkable robustness, as evidenced by its minimal performance degradation of only 0.71\% compared to H-SAM's 1.09\% and SAMed's 1.43\% drops, with this resilience likely stemming from OFL-SAM2's innovative online learning mechanism that enables dynamic adaptation to test sequences through continuous model refinement during inference. Some visualization comparisons of the multi-organ segmentation results are shown in the left part of Fig.\ref{vis_res}.

\begin{table}
  \centering
  \begin{tabular}{c|c|cc}
    \toprule
     \makecell{Training set} & \makecell{Method}& \makecell{Dice$ (\%)\uparrow$ }  &HD $\downarrow$\\
    \midrule
    None & SAM2 (mask) & 80.44 & 15.26 \\
    \midrule
    \multirow{6}{*}{\makecell{2 \\ volumes}}
    & SAMed &84.07 & 11.75 \\
    & SurgicalSAM & 84.93  & 10.62 \\
    & H-SAM & 85.44 & 9.30 \\
    \cmidrule{2-4}
    & FS-MedSAM2 & 82.89 & 12.64 \\
    & FATE-SAM2 & 83.43 & 12.16 \\
    & \textbf{OFL-SAM2} &\textbf{88.07}&\textbf{7.18}\\
    \midrule
    \multirow{6}{*}{\makecell{3 \\ volumes}}
    & SAMed & 86.12 & 10.57 \\
    & SurgicalSAM & 86.74  & 9.61 \\
    & H-SAM & 87.27 & 7.46 \\
    \cmidrule{2-4}
    & FS-MedSAM2 & 83.88 & 12.03 \\
    & FATE-SAM2 & 84.58 & 11.18 \\
    & \textbf{OFL-SAM2} &\textbf{88.74}&\textbf{6.49}\\
    \hline
  \end{tabular}
  \caption{Comparison results on the PROMISE12 dataset.}
  \label{tab:promise}
\end{table}

\begin{table}
  \centering
  \setlength{\tabcolsep}{1.1mm}
  \begin{tabular}{c|c|cc}
    \toprule
     \makecell{Training set} & \makecell{Method}& \makecell{Mean Dice (\%)$\uparrow$}  &HD $\downarrow$\\
    \midrule
    None & SAM2 (mask) & 74.63 & 19.02 \\
    \midrule
    \multirow{6}{*}{5\%}
    & SAMed & 76.36 & 17.64 \\
    & H-SAM & 80.22 & 14.48\\
    & SurgicalSAM & 80.76  & 13.32 \\
    \cmidrule{2-4}
    & FS-MedSAM2 & 78.94 & 15.12 \\
    & FATE-SAM2 &79.66 & 14.57 \\
    & OFL-SAM2 &\textbf{83.78}&\textbf{11.88}\\
    \midrule
    \multirow{6}{*}{10\%}
    & SAMed  & 78.83 & 15.46\\
    & H-SAM &82.80 & 12.76\\
    & SurgicalSAM & 83.04  &12.33 \\
    \cmidrule{2-4}
    & FS-MedSAM2 & 79.64 & 14.36 \\
    & FATE-SAM2 & 80.21 & 13.98\\
    & \textbf{OFL-SAM2} &\textbf{84.41}&\textbf{11.04}\\
    \hline
  \end{tabular}
  \caption{Comparison results on the Autolaparo dataset.}
  \label{tab:auto}
\end{table}

\paragraph{PROMISE12.} Here, we compare the performance of our proposed OFL-SAM2 and the comparison methods on the PROMISE12 dataset. As presented in Table~\ref{tab:promise}, without training, the SAM2 with a mask prompt attains 80.44\% DSC. When trained on 2 and 3 volumes of data, our model achieves the DSC of 88.07\% and 88.74\%, respectively, which outperforms the LoRA-fined SAM variants, such as H-SAM and SAMed, and the training-free FATE-SAM2. Under both the 2-volume and 3-volume training settings, OFL-SAM2 surpasses the second-place comparison method H-SAM by 2.63\% and 1.47\% in DSC, respectively, demonstrating the effectiveness of our model.

\paragraph{Autolaparo.} We also evaluate the performance of our approach on the surgical video dataset, with comparison results presented in Table~\ref{tab:auto}. When 10\% of the video sequence data is used for training, OFL-SAM2 achieves state-of-the-art performance of 84.41\% mean DSC and 11.04 HD, leading the second place SurgicalSAM by 1.37\% and the SAM2 variant FATE-SAM2 by 4.2\% on mean DSC. When only using 5\% of data, OFL-SAM2 still achieves the best accuracy of 83.78\% mean DSC, which is 3.56\% higher than H-SAM. Some visualizations of the instrument segmentation results are shown in the right part of Fig.\ref{vis_res}.

\begin{table}
  \centering
  \begin{tabular}{c|cc|c}
    \toprule
     \makecell{Dataset} & \makecell{Few-shot \\ learner} & AFM & \makecell{Mean Dice$\uparrow$\\(\%)}\\
    \midrule
    \multirow{3}{*}{Synapse-CT} 
    & $\times$ & $\times$ & 74.74 \\
    & \checkmark &  $\times$ & 78.96 \\
    & \checkmark &  \checkmark & \textbf{82.52}\\
    \hline
    \multirow{3}{*}{\makecell{PROMISE12}} 
    & $\times$ & $\times$ & 83.34\\
    & \checkmark & $\times$ & 86.21 \\
    & \checkmark & \checkmark & \textbf{88.74} \\
    \hline
  \end{tabular}
  \caption{Ablation study on the few-shot learner module and AFM on the Synapse-CT and PROMISE12 datasets.}
  \label{tab:module}
\end{table}

\begin{table}
  \centering
  \begin{tabular}{c|cc|c}
    \toprule
     \makecell{Dataset} & \makecell{Update \\ strategy} & $\gamma$ & \makecell{Mean Dice$\uparrow$\\(\%)} \\
    \midrule
    \multirow{4}{*}{Synapse-CT} 
    & $\times$ & - & 81.48 \\
    & \checkmark &  0.7 & 82.13 \\
    & \checkmark &  0.8 & \textbf{82.52}\\
    & \checkmark &  0.9 & 82.04\\
    \hline
    \multirow{4}{*}{\makecell{PROMISE12}} 
    & $\times$ & - & 88.17\\
    & \checkmark & 0.7 & 88.54 \\
    & \checkmark & 0.8 & \textbf{88.74} \\
    & \checkmark & 0.9 & 88.36\\
    \hline
  \end{tabular}
  \caption{Ablation study on the update strategy on the Synapse-CT and PROMISE12 datasets.}
  \label{tab:us}
\end{table}

\subsection{Model Analysis}
\noindent\textbf{Module ablation.} To verify the effectiveness of the proposed modules in our model, we perform ablation studies on the Synapse-CT and PROMISE12 datasets using 3 volumes of data. As shown in Table~\ref{tab:module}, the OFL-SAM2 without the few-shot learner and the AFM represents the model using only the memory attention module based on the most similar training image without training, which obtains 74.74\% mean DSC on the Synapse-CT dataset. When we add the few-shot learner, the dynamically updatable mapping network provides the model with discriminative target representations and generalization to test sequences, which significantly improves the mean DSC by 4.22\% to 78.96\%. Furthermore, when both the few-shot learner and the AFM are included, the model can better fuse the features of the two branches to adapt the SAM2 decoder, thus realizing a better 82.52\% mean DSC. These experiments prove the effectiveness of each module and also show that putting them together performs better.

\paragraph{Update strategy.} Here, we explore the effectiveness of the proposed update strategy and the optimal hyperparameter $\gamma$ selection within a certain range. We adopt the same data settings as module ablation and report the results in Table~\ref{tab:us}. It can be seen that there is a clear performance degradation of the model in the absence of the update strategy, whereas after using the strategy, the model achieves the optimal results on both datasets with $\gamma$ = 0.8. When $\gamma$ rises to 0.9, the accuracy drops. This might be that too high $\gamma$ instead inhibits the selection of high-quality predictions, resulting in ineffective updates of the mapping network and memory bank.

\begin{table}
  \centering
  \begin{tabular}{c|c|c}
    \toprule
     \makecell{Dataset} & Image number (K) & \makecell{Mean Dice$\uparrow$\\(\%)}\\
    \midrule
    \multirow{3}{*}{Synapse-CT} 
    & 1 & 81.86 \\
    & 2 & \textbf{82.52} \\
    & 3 & 82.21\\
    \midrule
    \multirow{3}{*}{\makecell{PROMISE12}} 
    & 1 & 88.27\\
    & 2 & \textbf{88.74} \\
    & 3 & 88.68 \\
    \hline
  \end{tabular}
  \caption{Ablation study on the number of most similar training images for memory bank.}
  \label{tab:number}
\end{table}

\paragraph{Number of most similar training images.} To investigate the influence of different numbers (K) of the most similar training images for the memory bank during inference, we conduct ablation experiments on the Synapse-CT and PROMISE12 datasets using 3 volumes of data, and the results are reported in Table \ref{tab:number}. It can be seen that when K=2, OFL-SAM2 achieves the optimal performance. When K=1, the model's performance decreases by about 0.7\% mean DSC on Synapse-CT and 0.5\% DSC on PROMISE12. The reason may be that a single training image struggles to provide enough target information. When K=3, the performance stabilizes and declines slightly, possibly due to saturation of the provided target information and a reduction of feature information for subsequent updates. Therefore, we select K=2 as the default setting for better accuracy.

\section{Conclusion}
In this work, we introduce OFL-SAM2, a prompt-free SAM2 framework for efficient MIS of both 3D volumes and temporally correlated 2D image sequences. To achieve prompt-free operation while providing discriminative target representations akin to prompts for each frame, we propose an online few-shot learner that trains a lightweight mapping network to capture medical knowledge and transforms generic image features into target features using limited data. Critically, the few-shot learner can update the mapping network parameters during inference, enhancing the model’s generalization across test sequences. Furthermore, to adapt the additional target representations to the frozen SAM2 framework and eliminate potential distraction representations, we devise an adaptive fusion module, which dynamically integrates the target features with the memory-attention features from SAM2, leading to accurate segmentations. Additionally, we propose a quality-aware update strategy that selectively incorporates high-confidence predictions for both the few-shot learner and memory bank updates to prevent error accumulation. Experimental results show that OFL-SAM2 achieves state-of-the-art performance on three diverse MIS datasets under data-limited conditions.

\section{Acknowledgements}
This work was partially supported by the National Natural Science Foundation of China (NSFC) (Grant No. 62306254); the Research Grants Council (RGC) of the Hong Kong Special Administrative Region, China (Project No. R6005-24); and the Hong Kong Joint Research Scheme (JRS) of the NSFC/RGC (Project No. N\_HKUST654/24).

\bibliography{main}

@String(CVPR= {IEEE Conf. Comput. Vis. Pattern Recog.})

@String(ICCV= {Int. Conf. Comput. Vis.})

@String(ECCV= {Eur. Conf. Comput. Vis.})

@String(BMVC= {Brit. Mach. Vis. Conf.})

@String(ICLR = {Int. Conf. Learn. Represent.})

@String(AAAI = {AAAI})

@String(CVPR  = {CVPR})

@String(ICCV  = {ICCV})

@String(ECCV  = {ECCV})

@String(BMVC  =	{BMVC})

@String(ICLR  = {ICLR})

@inproceedings{bhat2020learning,
  title={Learning what to learn for video object segmentation},
  author={Bhat, Goutam and Lawin, Felix J{\"a}remo and Danelljan, Martin and Robinson, Andreas and Felsberg, Michael and Van Gool, Luc and Timofte, Radu},
  booktitle={ECCV},
  pages={777--794},
  year={2020}
}

@inproceedings{landman2015miccai,
  title={Miccai multi-atlas labeling beyond the cranial vault--workshop and challenge},
  author={Landman, Bennett and Xu, Zhoubing and Igelsias, Juan and Styner, Martin and Langerak, Thomas and Klein, Arno},
  booktitle={MICCAI multi-atlas labeling beyond cranial vault—workshop challenge},
  volume={5},
  pages={12},
  year={2015}
}

@article{samed,
  title={Customized segment anything model for medical image segmentation},
  author={Zhang, Kaidong and Liu, Dong},
  journal={arXiv preprint arXiv:2304.13785},
  year={2023}
}

@article{he2025few,
  title={Few-Shot Adaptation of Training-Free Foundation Model for 3D Medical Image Segmentation},
  author={He, Xingxin and Hu, Yifan and Zhou, Zhaoye and Jarraya, Mohamed and Liu, Fang},
  journal={arXiv preprint arXiv:2501.09138},
  year={2025}
}

@inproceedings{hsam,
  title={Unleashing the potential of SAM for medical adaptation via hierarchical decoding},
  author={Cheng, Zhiheng and Wei, Qingyue and Zhu, Hongru and Wang, Yan and Qu, Liangqiong and Shao, Wei and Zhou, Yuyin},
  booktitle={CVPR},
  pages={3511--3522},
  year={2024}
}

@article{ma2024segment,
  title={Segment anything in medical images},
  author={Ma, Jun and He, Yuting and Li, Feifei and Han, Lin and You, Chenyu and Wang, Bo},
  journal={Nature Communications},
  volume={15},
  number={1},
  pages={654},
  year={2024}
}

@inproceedings{sam2023,
  title={Segment anything},
  author={Kirillov, Alexander and Mintun, Eric and Ravi, Nikhila and Mao, Hanzi and Rolland, Chloe and Gustafson, Laura and Xiao, Tete and Whitehead, Spencer and Berg, Alexander C and Lo, Wan-Yen and others},
  booktitle={ICCV},
  pages={4015--4026},
  year={2023}
}

@article{zhu2024medical,
  title={Medical sam 2: Segment medical images as video via segment anything model 2},
  author={Zhu, Jiayuan and Hamdi, Abdullah and Qi, Yunli and Jin, Yueming and Wu, Junde},
  journal={arXiv preprint arXiv:2408.00874},
  year={2024}
}

@inproceedings{wang2023sam,
author = {Wang, Haoyu and Guo, Sizheng and Ye, Jin and Deng, Zhongying and Cheng, Junlong and Li, Tianbin and Chen, Jianpin and Su, Yanzhou and Huang, Ziyan and Shen, Yiqing and Fu, Bin and Zhang, Shaoting and He, Junjun and Qiao, Yu},
title = {SAM-Med3D: Towards General-Purpose Segmentation Models for Volumetric Medical Images},
booktitle = {ECCV 2024 Workshops},
pages = {51–67},
year = {2025}
}

@article{wu2025medical,
  title={Medical sam adapter: Adapting segment anything model for medical image segmentation},
  author={Wu, Junde and Wang, Ziyue and Hong, Mingxuan and Ji, Wei and Fu, Huazhu and Xu, Yanwu and Xu, Min and Jin, Yueming},
  journal={Medical image analysis},
  volume={102},
  pages={103547},
  year={2025}
}

@article{sam2,
  title={Sam 2: Segment anything in images and videos},
  author={Ravi, Nikhila and Gabeur, Valentin and Hu, Yuan-Ting and Hu, Ronghang and Ryali, Chaitanya and Ma, Tengyu and Khedr, Haitham and R{\"a}dle, Roman and Rolland, Chloe and Gustafson, Laura and others},
  journal={arXiv preprint arXiv:2408.00714},
  year={2024}
}

@inproceedings{surgicalsam,
  title={Surgicalsam: Efficient class promptable surgical instrument segmentation},
  author={Yue, Wenxi and Zhang, Jing and Hu, Kun and Xia, Yong and Luo, Jiebo and Wang, Zhiyong},
  booktitle={AAAI},
  volume={38},
  number={7},
  pages={6890--6898},
  year={2024}
}

@inproceedings{yan2025pgp,
  title={PGP-SAM: Prototype-Guided Prompt Learning for Efficient Few-Shot Medical Image Segmentation},
  author={Yan, Zhonghao and Yin, Zijin and Lin, Tianyu and Zeng, Xiangzhu and Liang, Kongming and Ma, Zhanyu},
  booktitle={2025 IEEE 22nd International Symposium on Biomedical Imaging (ISBI)},
  pages={1--5},
  year={2025}
}

@inproceedings{xu2024sam,
title={{SAM}-{MPA}: Applying {SAM} to Few-shot Medical Image Segmentation using Mask Propagation and Auto-prompting},
author={Jie Xu and LiXiaokang and Chengyuyue and Chen Ma and Yi Guo and Yuanyuan Wang},
booktitle={Advancements In Medical Foundation Models: Explainability, Robustness, Security, and Beyond},
year={2024}
}

@article{revsam2,
    title={RevSAM2: Prompt SAM2 for Medical Image Segmentation via Reverse-Propagation without Fine-tuning}, 
    author={Yunhao Bai and Boxiang Yun and Zeli Chen and Qinji Yu and Yingda Xia and Yan Wang},
    journal={arXiv preprint arXiv:2409.04298},
    year={2024}
}

@inproceedings{lwl2020,
  title={Learning what to learn for video object segmentation},
  author={Bhat, Goutam and Lawin, Felix J{\"a}remo and Danelljan, Martin and Robinson, Andreas and Felsberg, Michael and Van Gool, Luc and Timofte, Radu},
  booktitle={ECCV},
  pages={777--794},
  year={2020}
}

@inproceedings{mao2021joint,
  title={Joint inductive and transductive learning for video object segmentation},
  author={Mao, Yunyao and Wang, Ning and Zhou, Wengang and Li, Houqiang},
  booktitle={ICCV},
  pages={9670--9679},
  year={2021}
}

@inproceedings{deng2023sam,
  title={Sam-u: Multi-box prompts triggered uncertainty estimation for reliable sam in medical image},
  author={Deng, Guoyao and Zou, Ke and Ren, Kai and Wang, Meng and Yuan, Xuedong and Ying, Sancong and Fu, Huazhu},
  booktitle={MICCAI},
  pages={368--377},
  year={2023}
}

@article{cheng2023sam,
  title={Sam on medical images: A comprehensive study on three prompt modes},
  author={Cheng, Dongjie and Qin, Ziyuan and Jiang, Zekun and Zhang, Shaoting and Lao, Qicheng and Li, Kang},
  journal={arXiv preprint arXiv:2305.00035},
  year={2023}
}

@article{roy2023sam,
  title={Sam. md: Zero-shot medical image segmentation capabilities of the segment anything model},
  author={Roy, Saikat and Wald, Tassilo and Koehler, Gregor and Rokuss, Maximilian R and Disch, Nico and Holzschuh, Julius and Zimmerer, David and Maier-Hein, Klaus H},
  journal={arXiv preprint arXiv:2304.05396},
  year={2023}
}

@article{hu2022lora,
  title={Lora: Low-rank adaptation of large language models.},
  author={Hu, Edward J and Shen, Yelong and Wallis, Phillip and Allen-Zhu, Zeyuan and Li, Yuanzhi and Wang, Shean and Wang, Lu and Chen, Weizhu and others},
  journal={ICLR},
  volume={1},
  number={2},
  pages={3},
  year={2022}
}

@inproceedings{
chen2024sam2,
title={{SAM}2-Adapter: Evaluating \& Adapting Segment Anything 2 in Downstream Tasks: Camouflage, Shadow, Medical Image Segmentation, and More},
author={Tianrun Chen and Ankang Lu and Lanyun Zhu and Chaotao Ding and Chunan Yu and Deyi Ji and Zejian Li and Lingyun Sun and Papa Mao and Ying Zang},
booktitle={ICLR 2025 Workshop on Foundation Models in the Wild},
year={2025}
}

@inproceedings{xiao2024cat,
  title={Cat-sam: Conditional tuning for few-shot adaptation of segment anything model},
  author={Xiao, Aoran and Xuan, Weihao and Qi, Heli and Xing, Yun and Ren, Ruijie and Zhang, Xiaoqin and Shao, Ling and Lu, Shijian},
  booktitle={ECCV},
  pages={189--206},
  year={2024}
}

@inproceedings{lin2024beyond,
  title={Beyond adapting SAM: Towards end-to-end ultrasound image segmentation via auto prompting},
  author={Lin, Xian and Xiang, Yangyang and Yu, Li and Yan, Zengqiang},
  booktitle={MICCAI},
  pages={24--34},
  year={2024}
}

@article{litjens2014evaluation,
  title={Evaluation of prostate segmentation algorithms for MRI: the PROMISE12 challenge},
  author={Litjens, Geert and Toth, Robert and Van De Ven, Wendy and Hoeks, Caroline and Kerkstra, Sjoerd and Van Ginneken, Bram and Vincent, Graham and Guillard, Gwenael and Birbeck, Neil and Zhang, Jindang and others},
  journal={Medical Image Analysis},
  volume={18},
  number={2},
  pages={359--373},
  year={2014}
}

@inproceedings{wang2021transformer,
  title={Transformer meets tracker: Exploiting temporal context for robust visual tracking},
  author={Wang, Ning and Zhou, Wengang and Wang, Jie and Li, Houqiang},
  booktitle={CVPR},
  pages={1571--1580},
  year={2021}
}

@inproceedings{danelljan2020,
  title={Probabilistic regression for visual tracking},
  author={Danelljan, Martin and Gool, Luc Van and Timofte, Radu},
  booktitle={CVPR},
  pages={7183--7192},
  year={2020}
}

@inproceedings{robinson2020,
  title={Learning fast and robust target models for video object segmentation},
  author={Robinson, Andreas and Lawin, Felix Jaremo and Danelljan, Martin and Khan, Fahad Shahbaz and Felsberg, Michael},
  booktitle={CVPR},
  pages={7406--7415},
  year={2020}
}

@inproceedings{song2023meta,
  title={Meta-adapter: An online few-shot learner for vision-language model},
  author={Song, Lin and Xue, Ruoyi and Wang, Hang and Sun, Hongbin and Ge, Yixiao and Shan, Ying and others},
  booktitle={NeurIPS},
  volume={36},
  pages={55361--55374},
  year={2023}
}

@inproceedings{autolaparo,
  title={Autolaparo: A new dataset of integrated multi-tasks for image-guided surgical automation in laparoscopic hysterectomy},
  author={Wang, Ziyi and Lu, Bo and Long, Yonghao and Zhong, Fangxun and Cheung, Tak-Hong and Dou, Qi and Liu, Yunhui},
  booktitle={MICCAI},
  pages={486--496},
  year={2022}
}

@inproceedings{adamW,
  title={Decoupled Weight Decay Regularization},
  author={Loshchilov, Ilya and Hutter, Frank},
  booktitle={International Conference on Learning Representations},
  year={2019}
}

@article{bai2024fs,
  title={Fs-medsam2: Exploring the potential of sam2 for few-shot medical image segmentation without fine-tuning},
  author={Bai, Yunhao and Yu, Qinji and Yun, Boxiang and Jin, Dakai and Xia, Yingda and Wang, Yan},
  journal={arXiv e-prints},
  pages={arXiv--2409},
  year={2024}
}

@article{ma2025medsam2,
  title={Medsam2: Segment anything in 3d medical images and videos},
  author={Ma, Jun and Yang, Zongxin and Kim, Sumin and Chen, Bihui and Baharoon, Mohammed and Fallahpour, Adibvafa and Asakereh, Reza and Lyu, Hongwei and Wang, Bo},
  journal={arXiv preprint arXiv:2504.03600},
  year={2025}
}

@ARTICLE{huai2025tmi,
  author={Huai, Zheang and Tang, Hui and Li, Yi and Chen, Zhuangzhuang and Li, Xiaomeng},
  journal={IEEE Transactions on Medical Imaging}, 
  title={Leveraging Segment Anything Model for Source-Free Domain Adaptation via Dual Feature Guided Auto-Prompting}, 
  year={2025}
}

@article{zhang2024glandsam,
  title={GlandSAM: injecting morphology knowledge into segment anything model for label-free gland segmentation},
  author={Zhang, Qixiang and Li, Yi and Xue, Cheng and Wang, Haonan and Li, Xiaomeng},
  journal={IEEE Transactions on Medical Imaging},
  year={2024}
}

@inproceedings{wang2024tri,
  title={Tri-plane mamba: Efficiently adapting segment anything model for 3d medical images},
  author={Wang, Hualiang and Lin, Yiqun and Ding, Xinpeng and Li, Xiaomeng},
  booktitle={International Conference on Medical Image Computing and Computer-Assisted Intervention},
  pages={636--646},
  year={2024},
  organization={Springer}
}

@article{li2018h,
  title={H-DenseUNet: hybrid densely connected UNet for liver and tumor segmentation from CT volumes},
  author={Li, Xiaomeng and Chen, Hao and Qi, Xiaojuan and Dou, Qi and Fu, Chi-Wing and Heng, Pheng-Ann},
  journal={IEEE transactions on medical imaging},
  volume={37},
  number={12},
  pages={2663--2674},
  year={2018},
  publisher={IEEE}
}

@article{li2020transformation,
  title={Transformation-consistent self-ensembling model for semisupervised medical image segmentation},
  author={Li, Xiaomeng and Yu, Lequan and Chen, Hao and Fu, Chi-Wing and Xing, Lei and Heng, Pheng-Ann},
  journal={IEEE transactions on neural networks and learning systems},
  volume={32},
  number={2},
  pages={523--534},
  year={2020}
}

@inproceedings{li2018semi,
  title={Semi-supervised Skin Lesion Segmentation via Transformation Consistent Self-ensembling Model},
  author={Li, Xiaomeng and Yu, Lequan and Chen, Hao and Fu, Chi Wing and Heng, Pheng Ann},
  booktitle={British Machine Vision Conference (BMVC)},
  year={2018}
}

\end{document}